\documentclass[10pt,twocolumn,letterpaper]{article}

\usepackage{cvpr}
\usepackage{times}
\usepackage{epsfig}
\usepackage{graphicx}
\usepackage{amsmath}
\usepackage{amssymb}
\usepackage{caption}
\usepackage{subcaption}

\graphicspath{{figs/pdf/}}

\usepackage[pagebackref=true,breaklinks=true,letterpaper=true,colorlinks,bookmarks=false]{hyperref}
\usepackage{cleveref}
\crefname{subfigure}{fig.}{fig.}

\def\version{final}

\usepackage[pdftex,usenames]{color}
\def\textred#1{\textcolor{red}{#1}}
\def\textgrn#1{\textcolor{green}{#1}}
\def\textblu#1{\textcolor{blue}{#1}}

\usepackage{ifthen}

\def\paperid{600}
\def\cvprPaperID{\paperid} 

\ifthenelse{\equal{\version}{draft}}{
   \def\myrnote#1#2{{\sl\textred{\small[#1:#2]}}}
   \def\mygnote#1#2{{\sl\textgrn{\small[#1:#2]}}}
   \def\mybnote#1#2{{\sl\textblu{\small[#1:#2]}}}
    \def\mgnote#1{\mygnote{MG}{#1}}
    \def\hrnote#1{\mybnote{HR}{#1}}
    \def\ienote#1{\myrnote{IE}{#1}}
    \def\HUSSAIN#1{\mybnote{HR}{#1}}
    \def\MATTHIAS#1{\mygnote{MG}{#1}}
    \def\IRFAN#1{\myrnote{IE}{#1}}    
    \def\cvprPaperID{\textred{\paperid, DRAFT}}
    }{
        \def\mygnote#1#2{} \def\myrnote#1#2{} \def\mybnote#1#2{}
        \def\ienote#1{}  \def\mgnote#1{} \def\hrnote#1{}
        \def\MATTHIAS#1{} \def\HUSSAIN#1{} \def\IRFAN#1{}
        \def\cvprPaperID{\paperid}
        \typeout{*************In Final MODE}
        }

\def\eg{\textit{e.g.~}}
\def\ie{\textit{i.e.~}}
\def\etc{\textit{etc}}
\def\etal{\textit{et~al.~}}

 \cvprfinalcopy 

\ifcvprfinal\fi
\begin{document}

\title{\vspace*{-0.15in}Geometric Context from Videos}

\author{
S. Hussain Raza
\hspace{10mm}
Matthias Grundmann
\hspace{10mm}
Irfan Essa
\\
{\small Georgia Institute of Technology, Atlanta, GA, USA}
\\
\centering
 {\href{http://www.cc.gatech.edu/cpl/projects/videogeometriccontext}{\small http://www.cc.gatech.edu/cpl/projects/videogeometriccontext}}
}

\maketitle

\begin{abstract}
\noindent We present a novel algorithm for estimating the broad 3D geometric structure of outdoor \emph{video} scenes. Leveraging \emph{spatio-temporal} video segmentation, we decompose a dynamic scene captured by a video into geometric classes, based on predictions made by region-classifiers that are trained on appearance and motion features. By examining the homogeneity of the prediction, we combine predictions across \emph{multiple} segmentation hierarchy \emph{levels} alleviating the need to determine the granularity a priori. We built a novel, extensive dataset on geometric context of video to evaluate our method, consisting of over 100 \emph{ground-truth annotated} outdoor videos with over 20,000 frames. To further scale beyond this dataset, we propose a semi-supervised learning framework to expand the pool of labeled data with high confidence predictions obtained from unlabeled data. Our system produces an accurate prediction of geometric context of video achieving 96\% accuracy across main geometric classes.
\end{abstract}


\section{Introduction}
\label{sec:intro}

\noindent Holistic scene understanding requires an understanding of the broad 3D structure of the scene with all objects present. One important step towards this goal is to partition a scene into regions and label them relative to each other and within the scene geometry. Geometric classes can define the basic 3D structure of a scene with respect to the camera, and suggest cues to identify horizontal surfaces and vertical objects in the scene. Hoeim~\etal\cite{hoiem2006putting} showed that such geometric context can be used to obtain a probabilistic representation of the scene layout based on geometric classes, which in turn can be used to improve object detection. Torralba~\etal \cite{torralba2004contextual} showed that global context plays an important role in object detection. Recently, Divala~\etal\cite{empiricalContext} showed that incorporating geometric context, not only improves object detection but also makes misclassifications more reasonable. 

\begin{figure}
\centering
\includegraphics[scale=.31]{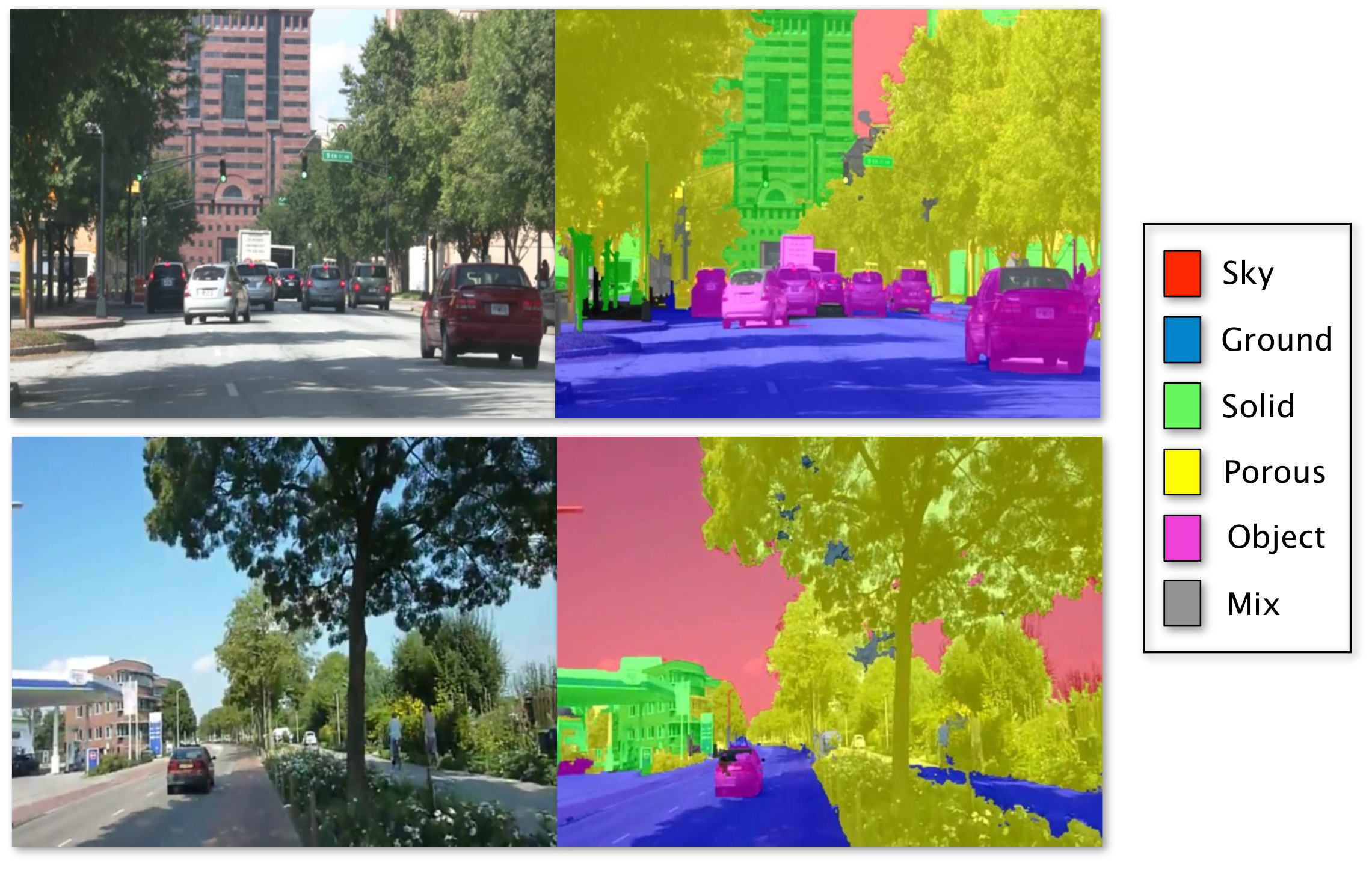}
\vspace{-0.12in}
\caption{\small{Video frames of an urban scene (left) and the predicted geometric context of our system (right). We achieve high accuracy leveraging motion and appearance features while achieving temporal consistency by relying on spatio-temporal regions across various granularities. Please watch the accompanying video.}}
\vspace{-0.12in}
\label{fig:f1}
\end{figure}

In this paper, we propose a novel method to provide a high level description of a \emph{video} scene by assigning geometric classes to spatio-temporal regions as shown in \Cref{fig:f1}. Building upon a hierarchical video-segmentation to achieve temporal consistency, we compute a wide variety of appearance, location, and motion features which are used to train classifiers to predict geometric context in video. A significant challenge for developing scene understanding system for videos is a need for an annotated \emph{video} dataset available for training and evaluation. To this end, we have collected and annotated a video dataset with pixel level ground truth labels for over 20,000 frames across 100 videos covering a wide variety of scene examples. 

The primary contributions of this paper are:
\begin{list}{$\bullet$}
{\setlength{\leftmargin}{1ex}\setlength{\labelsep}{.5ex}\setlength{\parsep}{.4ex plus .2ex minus .1ex}\setlength{\itemsep}{0ex plus .1ex}\setlength{\topsep}{0ex plus .1ex}}
\item A scene description for video via geometric classes (96\% accuracy across main geometric classes).
\item Exploiting motion and temporal causality/redundancy present in video by using motion features and aggregating predictions across spatio-temporal regions.
\item A semi-supervised bootstrap learning framework for expanding the pool of labeled data with highly confident predictions obtained on unlabeled data.
\item A novel dataset of 100 annotated videos ($\sim$20,000 frames) with pixel level labels, which will be made available. 
\item A thorough evaluation of our system by examining importance of features, benefit of temporal redundancy and independence of segmentation granularity.
\end{list}

\section{Related Work}

\noindent Image based scene understanding methods\cite{HoiemIJCV2007,gould2009decomposing} can be directly applied to individual video frames to generate a description of the scene. However, these methods do not exploit temporal information across neighboring frames. Further, lacking temporal consistency, they can result in temporally inconsistent labels across frames, which can impact performance, as scene labels suddenly change between frames. In addition, frame-based methods do not exploit temporal redundancy to process videos efficiently as processing each segment in video independently results in a longer processing time. 

Temporal information can be incorporated using structure from motion (SfM)~\cite{sturgess2009combining,brostow2008segmentation}, which requires substantial computation and might not generalize well to dynamic objects. SfM makes explicit assumptions about the scene, \eg, mostly static with limited foreground motion, and requires basic camera priors. In addition, SfM reconstruction can fail due to lack of parallax, \eg, walking forward, and rotation around camera center. We do not make any assumption about the scene content, amount of foreground motion, or the camera used. Another approach to achieve temporal consistency across frames is to use optical flow between consecutive frames to estimate the neighborhood of each pixel and then combine past predictions to make a final prediction~\cite{miksik-cmutr-12}. This requires labeling every pixel in every frame in the video independently, which doesn't leverage the causality in video. 

Our video scene understanding approach takes advantage of spatio-temporal information  by employing hierarchical video segmentation\cite{MatthiasSegmentation}, which segments a video into spatio-temporal regions. Further, we leverage causality in videos to efficiently label videos, achieving favorable complexity which is linear in the number of unique \emph{spatio-temporal} segments in videos. Consequently, in contrast to image based or independent frame labeling, our system is not directly affected by the total number of frames. Recently, Tighe \etal\cite{tighe2012superparsing} applied their image label transfer to the video domain leveraging\cite{MatthiasSegmentation}, by applying a max heuristic across frames. In contrast, our approach performs geometric labeling by leveraging multiple hierarchy level while probabilistic aggregating labels  over a temporal window.

A significant hurdle in video scene understanding is the availability of a ground truth annotated dataset for training. While several datasets exist for predicting geometric context in the \emph{image} domain \cite{HoiemIJCV2007, gould2009decomposing}, datasets for videos \cite{brostow2009semantic, NYUscene, MPIVehData} are currently limited in their scope. (see \cref{sec:classes}).

Our video scene analysis method builds upon Hoeim \etal's \cite{HoiemIJCV2007} image based approach, extending the image based approach to video. Our approach differs, in that it is taking advantage of spatio-temporal context, extends feature set being more suitable for video, leverages temporal redundancy while achieving temporal consistency and broadens the pool of available data by semi-supervised learning.
\section{Dataset and Geometric Classes}
\label{sec:classes}

\noindent \textbf{Existing Datasets:} 
In our supervised learning setting, we require an annotated dataset supplying ground truth labels for training and evaluation. While several datasets for geometric scene understanding exists on still images~\cite{HoiemIJCV2007, gould2009decomposing}, our video-based scene analysis method demands an annotated video dataset. However, existing datasets for video scene understanding only provide limited ground truth data. The CamVid dataset \cite{brostow2009semantic} provides pixel-level labels for 701 non-consecutive frames (about every 30th frame, sampled at 1fps). The NYUScenes \cite{NYUscene}, and MPI-VehicleScenes \cite{MPIVehData} dataset consists of 74 and 156 annotated frames, respectively. Therefore, these datasets are not ideally suited for comprehensive studies. To overcome this limitation, we provide a novel,  pixel-level annotated dataset for geometric scene analysis of video, consisting of over 20,000 frames across 100 videos.
\vspace{-0.18in}
\paragraph{A video dataset for geometric scene understanding:} 
Our dataset consists of 160 outdoor videos, with annotations available for a subset of 100 videos. Some videos are collected from YouTube and others are recorded by us while walking or driving in an urban area. Video lengths range from 60 to 400 frames and resolution varies from $320\times480$ to $600\times800$, with varying aspect ratios. We partitioned the datasets into three sets: 63 videos are used for training and cross-validation (13,000 frames), 40 videos for independent testing via external-validation (7,000 frames), and 60 videos are kept unlabeled (14,000 frames) and are later used for semi-supervised learning (\Cref{sec:semisupervised}). Videos in the cross and external-validation sets are completely annotated with ground truth labels (every frame and pixel).

Videos in our dataset contain entities such as sky, ground, buildings, trees, and objects (cars, trains, humans). While many different partitions can be imagined, we follow \cite{HoiemIJCV2007, HoiemGeometric2005} and partition the video content into three main geometric classes: ``Sky'', ``support'', and ``vertical''. 
To provide a more detailed description of the scene, we further divide the vertical class into three subclasses: ``Solid'', ``porous'', and ``object''. The solid vertical sub-class includes \emph{solid, static} objects resting on the ground, such as buildings, boards, bridges, and rocks etc. The porous vertical sub-class includes \emph{non-solid, static} objects such as trees and foliage. Finally, \emph{movable} objects, like humans, cars, boats, and trains are included in the object class. Notice, that in contrast to \cite{HoiemIJCV2007, HoiemGeometric2005} we do not account for the orientation of the vertical classes as their identity is likely to change due to camera motion in video. \Cref{Table:pix_area} gives an overview of the distribution of the classes in the cross-validation dataset, by showing the pixel area of each of the geometric classes. 
\begin{table}
\centering
\begin{subtable}[b]{0.45\columnwidth}
\small
\centering
\begin{tabular}{|c|c|}\hline
\emph{Sky} & 32.5\% \\\hline
\emph{Ground} & 26.4\%  \\\hline
\emph{Vertical} & 40.6\% \\\hline
\emph{mix} & 0.5\%   \\\hline
\end{tabular}
\vspace*{-0.05in}
\caption{Main Classes}
\end{subtable}
\begin{subtable}[b]{0.45\columnwidth}
\small
\centering
\begin{tabular}{|c|c|}\hline
\emph{Solid} & 19.7\%  \\\hline
\emph{Porous} &  15.6\%  \\\hline
\emph{Object}  & 3.7\%  \\\hline
\end{tabular}
\vspace*{-0.05in}
\caption{Sub-vertical Classes}
\end{subtable}
\vspace*{-0.1in}
\caption{\small{Average area in pixels of each geometric class in the cross-validation dataset.}}
\label{Table:pix_area}
\vspace*{-0.1in}
\end{table}

\section{Geometric Context From Videos}

\begin{figure}[!t]
\centering
\includegraphics[scale=.32]{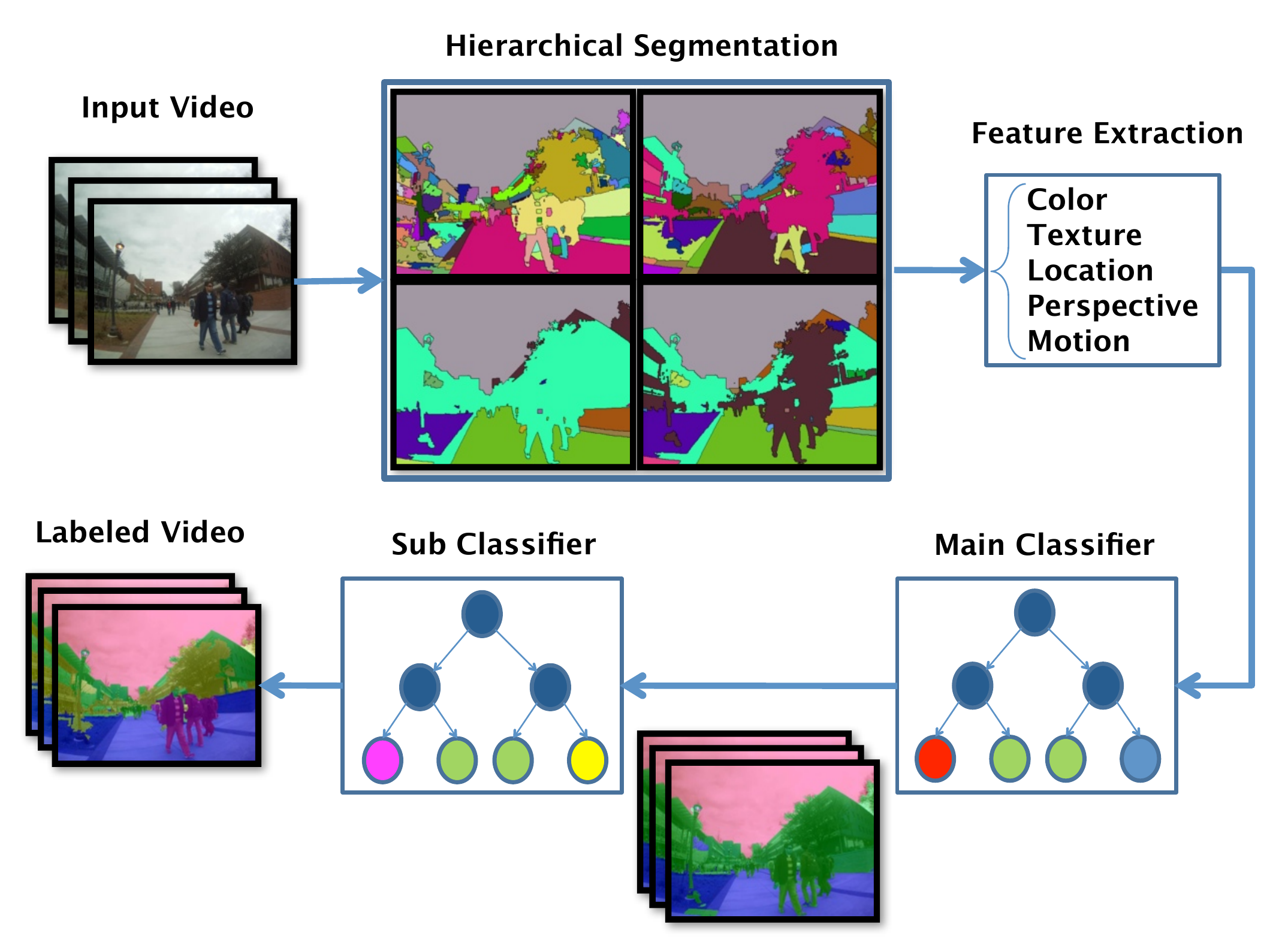}
\vspace*{-0.1in}
\caption{\small{Overview of our approach. First, input videos are segmented into a hierarchy of spatio-temporal regions using \cite{MatthiasSegmentation}. Then, features are extracted for each segment to train a main and sub-classifier to predict geometric context in videos.}}
\vspace*{-0.1in}
\label{fig:flow}
\end{figure}

\noindent Our algorithm for determining geometric context from video consists of 3 main steps (\Cref{fig:flow}). First, we apply hierarchical video segmentation, obtaining spatio-temporal regions at different hierarchy levels. We rely on video segmentation to achieve (a) temporal coherence without having to explicitly enforce it in our framework and (b) by labeling regions as opposed to individual pixels we greatly reduce computational complexity. Second, we extract several features from each segment. Third, we train a classifier to discriminate segments into \emph{sky}, \emph{ground}, and \emph{vertical} classes. Additionally, a sub-classifier is trained to discriminate the vertical class further into \emph{solid}, \emph{porous}, and \emph{object}. In particular, we employ a boosted decision tree classifier with a logistic version of Adaboost~\cite{LogisticAdaboost}. We will describe each of the above steps in more detail below.

\subsection{Video Segmentation}
\label{sec:video_segmentation}
\begin{figure}[!t]
\centering
\includegraphics[width=0.26\columnwidth]{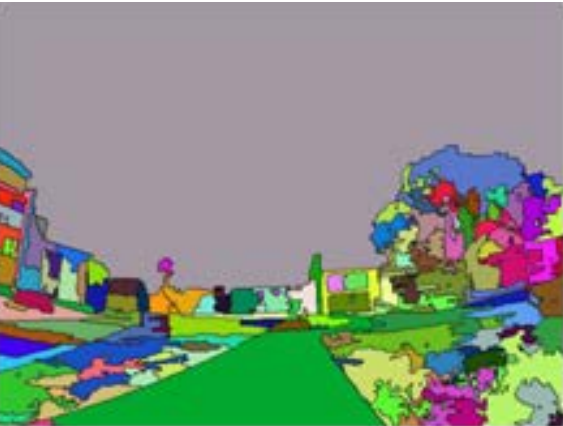}
\includegraphics[width=0.26\columnwidth]{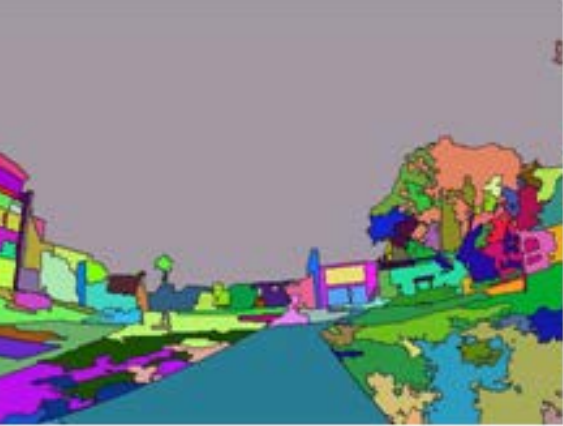}
\includegraphics[width=0.26\columnwidth]{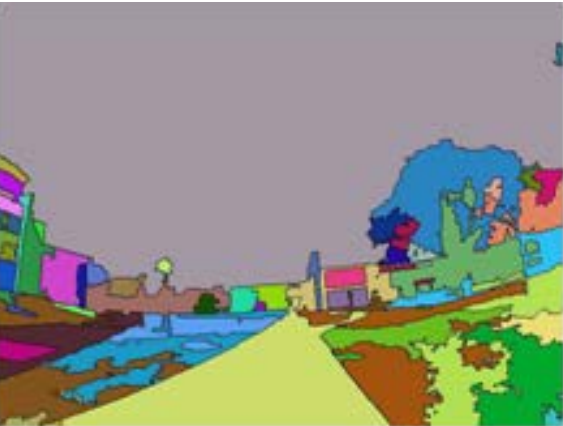}
\caption{\small{Video segmentation results by applying \cite{MatthiasSegmentation}. From left to right: Hierarchy levels in increasing order; region area increases as segments from lower hierarchy levels are grouped together.}}
\vspace*{-0.15in}
\label{fig:hie}
\end{figure}

\noindent Video segmentation aims to group similar pixels into spatio-temporal regions that are coherent in both appearance and motion. We use the hierarchical graph-based video segmentation algorithm proposed by Grundmann \etal\cite{MatthiasSegmentation,Corso2012Evaluation}, which is automatic and achieves long-term coherence. For completeness, we give a brief overview of their algorithm. Their spatio-temporal hierarchical video segmentation builds upon the graph-based image segmentation of Felzenszwalb \etal\cite{superpixel} by constructing a graph over the 3-d space-time neighbors of a voxel. This approach generates an \emph{over}-segmented video volume, which is further segmented into a hierarchy of super-regions of varying granularity. After computing region descriptors based on appearance and motion, a graph is constructed where each region from the over-segmentation forms a node and is connected to its incident regions by an edge with a weight equal to the $\chi^2$-difference of their local descriptors. This so-called region graph is used to group the over-segmented regions into super-regions by applying \cite{superpixel} to the graph. Successive application of this algorithm yields a segmentation hierarchy of the video as shown in \Cref{fig:hie} for one of our sample videos.

\subsection{Video Annotation}
\label{sec:annotation}
\begin{figure}[!h]
\centering
\includegraphics[scale=.28,trim=0 0 0 1cm]{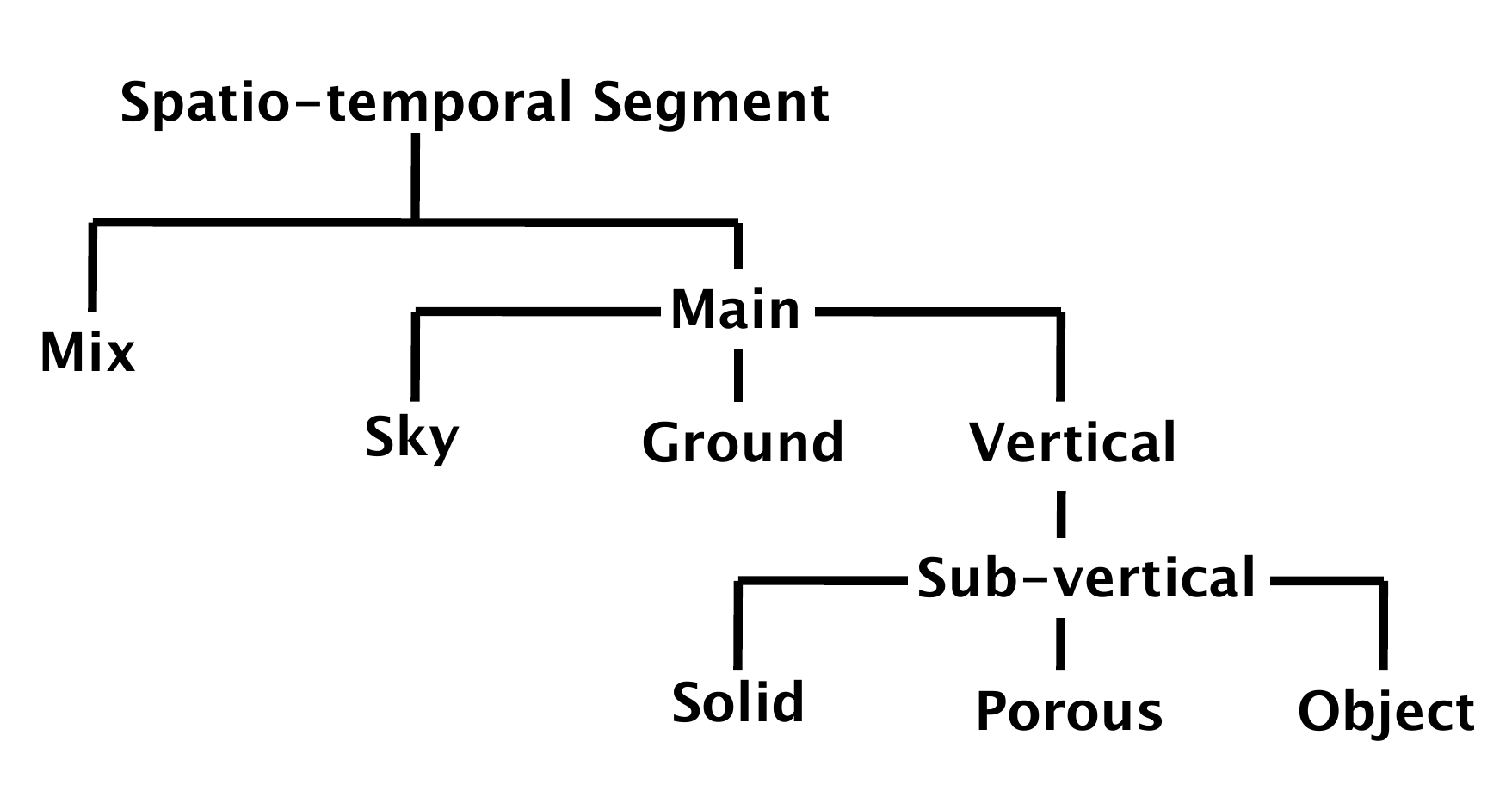}
\vspace*{-0.15in}
\caption{\small{Annotation hierarchy of spatio-temporal segments: Segments are either labeled as either as a mixture of classes (mix) or assigned a main geometric class label. The vertical geometric class is further discriminated into solid, porous, and object. }}
\vspace{-0.1in}
\label{fig:anno_hierachy}
\end{figure}

\noindent To obtain the ground truth for training and evaluation, we manually annotate over 100 videos. To greatly speed up the labeling process, we assign labels to individual spatio-temporal regions as opposed to pixels. In particular, we leverage the over-segmentation (\Cref{sec:video_segmentation}) to assign the appropriate label to each supervoxel. Though errors in the over-segmentation are limited due to the fine granularity, we need to address potential \emph{under-segmentation} errors, \ie a supervoxel contains more than one class. This is particularly of concern for the vertical class which contains a wide variety of potentially overlapping surfaces, \eg, buildings and trees, or several moving objects as cars, boats, trains, \etc. To address this problem, we introduce a new label ``mix'' to label a super-voxel, which is a mixture of two or more classes or if its identity is changing over time across geometric classes. \Cref{fig:anno_hierachy} shows the labels and their hierarchical relationship.

\begin{figure}[!t]
\centering
\includegraphics[scale=0.18]{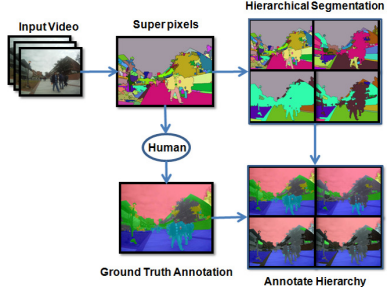}
\vspace*{-0.1in}
\caption{\small{Video annotation for obtaining ground truth: Over-segmented super-voxels are annotated manually. Supervoxel labels are then combined to generate ground truth for each level of segmentation hierarchy (see \Cref{sec:annotation}).
}}
\vspace{-0.05in}
\label{fig:anno}
\end{figure}

To obtain a ground truth labeling for every level of the segmentation hierarchy, we leverage the ground truth labels of the over-segmented super-voxels, by pooling their labels across a super-region via majority voting (a super-region is composed of several super-voxels). Specifically, if more than 95\% of a super-region's area is assigned the same ground truth label $L$ (based on the over-segmented super-voxels it is comprised of), the super-region is assigned label $L$, otherwise it is labeled as ``mix'', as shown in \Cref{fig:anno}. We manually annotated over 20,000 frames at the over-segmentation level and then combined their labels via the above approach across the hierarchy to generate labels at higher levels. \Cref{Table:segments} gives an overview of the percentage of segments annotated for each geometric class.

\begin{table}[Ht]
\centering
\begin{subtable}[b]{0.45\columnwidth}
\small
\centering
\begin{tabular}{|c|c|c|c|}\hline
\emph{Sky}& 2.5\% \\\hline
\emph{Ground} & 15.9\%  \\\hline
\emph{Vertical}&  81.2\% \\\hline
\emph{Mix}  & 0.4\%   \\\hline
\end{tabular}
\vspace{-0.05in}
\caption{Main Classes}
\end{subtable}
\begin{subtable}[b]{0.45\columnwidth}
\small
\centering
\begin{tabular}{|c|c|c|}\hline
\emph{Solid} & 47.5\% \\\hline
\emph{Porous}& 26.1\% \\\hline
\emph{Object}  & 7.7\%  \\\hline
\end{tabular}
\vspace{-0.05in}
\caption{Sub-vertical Classes}
\end{subtable}
\vspace{-0.1in}
\caption{\small{Percentage of segments annotated for each geometric class ($\sim2.5M$ in total at over segmented base hierarchy level).}}
\label{Table:segments}
\vspace{-0.15in}
\end{table}

\subsection{Features}
\label{sec:features}
\noindent We estimate the class-dependent probability of each geometric label for a segment in a frame using a wide variety of features. By segments, we refer to 2D per frame regions of the 3D spatio-temporal voxels. Specifically, we compute appearance (color, texture, location, perspective) and motion features \emph{across} each segment in a frame. 
For computing appearance features, we follow Hoeim \etal\cite{HoiemIJCV2007} and apply the publicly available code on a per-frame basis. For details please refer to \cite{HoiemIJCV2007}.

In videos, an additional feature not found in images is motion across frames. For motion features, we compute a histogram of dense optical flow (using OpenCV's implementation of Farneback's algorithm~\cite{farneback2003two}) as well as the mean motion of a segment. To capture the motion and changes in velocity and acceleration of objects across time, we compute flow histograms and mean flow for each frame $I_j$ \wrt to 3 previous frames: $I_{j-1}, I_{j-3}, I_{j-5}$. In particular, a segment $S_k$ we compute a 16-bin histogram of oriented flow vectors weighted by their corresponding magnitude. Histograms are normalized by the region area of the segment in current frame. In addition, we compute histograms for spatial flow differentials in $x$ and $y$, \ie  for the dense optical flow field $O = [O_x, O_y]$, we compute [$\partial_{x}O_x, \partial_{x}O_y]$ and $[\partial_{y}O_x, \partial_{y}O_y]$. To account for different scales, the flow differentials are computed for different kernel size of the Sobel filter (3, 5 and 7). As with the flow histograms, the spatial flow differentials are computed \wrt to 3 previous frames: $I_{j-1}, I_{j-3}, I_{j-5}$. This is similar to the approach of \cite{dalal2006flow}, which has been shown to to be helpful to the task of object detection in videos. \Cref{Table:feat} lists all of our motion based features used for estimating geometric context of video. 

\begin{table}[!h]
\centering
\small
\begin{tabular}{|c l|}
  \hline
  \multicolumn{2}{|c|}{\textbf{Motion based Features}} \\
  \hline
  \textbf{Dimensions} & \textbf{Description} \\
  $16\times3$ & Histogram of dense optical flow O of \\
	  &reference frame $I_j$ \wrt $I_{j-1}, I_{j-3}, I_{j-5}$.\\
  $16\times2\times3\times3$ & Histogram of differential of dense \\
	  &optical flow O in x and y, \\
	  & \ie [$\partial_{x}O_x, \partial_{x}O_y]$ and $[\partial_{y}O_x, \partial_{y}O_y$], across\\
	  & 3 kernel sizes of differential (3, 5, and 7) \\
	  & for reference frame $I_j$ \wrt $I_{j-1}, I_{j-3}, I_{j-5}$.\\
  $2\times3$ & Mean flow of a segment minus min. mean\\
  &  flow across all segments of current frame.\\
  $2\times3$ & Mean location change in x and y \\
  & for reference frame $I_j$ \wrt $I_{j-1}, I_{j-3}, I_{j-5}$.\\
  $2 \times 2 \times 3$ & 10th and 90th percentile of location change\\
  	& in x and y for frame $I_j$  \wrt $I_{j-3}, I_{j-5}$.\\\
  $1 \times 3$ & Magnitude of location change of mean,\\
  	& 10th and 90th percentile. \\  
  \hline
\end{tabular}
\vspace{-0.1in}
\caption{\small{List of flow and motion based features computed per frame $I_j$ and per segment $S_k$. See text for details. Appearance features are adopted using the approach of Hoeim \etal~\cite{HoiemIJCV2007}.}}
\label{Table:feat}
\vspace{-0.3in}
\end{table}

\subsection{Multiple Segmentations}
\noindent As the appropriate granularity of the segmentation is not known a priori, we make use of multiple segmentations across several hierarchy levels, utilizing the increased spatial support of the segments at higher levels to compute features. In particular, we combine the individually predicted labels based on homogeneity of the segments. Homogeneity is defined in our case as the probability of the segment \emph{not} being a mixture of several classes, \ie \emph{not} having the label mix (for details, see \Cref{sec:classification}). We generate multiple segmentations of the scene at various granularity levels ranging from 10\% to 50\% of the hierarchy height using \cite{MatthiasSegmentation} in increments of 10\% (5 hierarchy levels in total).

\begin{figure*}
\centering
\includegraphics[width=0.7\textwidth]{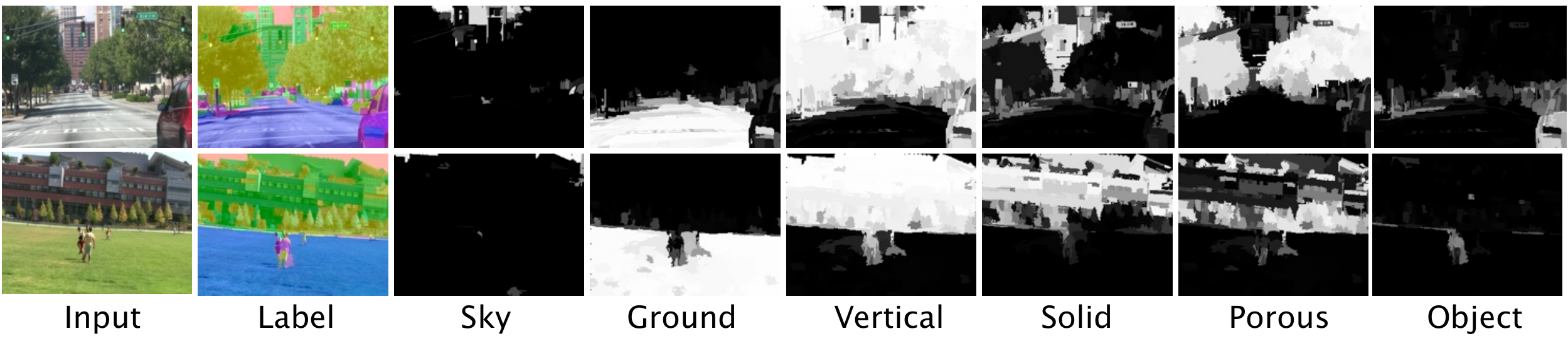}
\vspace{-0.1in}
\caption{\small{Input video image, predicted labels and confidence for each geometric class. Notice, that trees are correctly assigned high probability for porous class, walls for solid class, and humans and cars have high confidence for the object class.}}
\vspace{-0.1in}
\label{fig:vert_conf}
\end{figure*}

\subsection{Classification}
\label{sec:classification}
\noindent We evaluate our method using boosted decision trees based on a logistic regression version of Adaboost~\cite{LogisticAdaboost} that outputs the class probability for each segment in a frame and perform 5-fold cross validation. We train two multi-class classifiers to predict the geometric labels, first to discriminate among the main classes (sky, ground, and vertical), and second for further sub-classification of vertical class (buildings, porous, and objects). In addition to the two multi-class classifiers, we independently train a homogeneity classifier that estimates the probability of the segment being a single label segment or part of the class ``mix''. We refer to this probability as the \emph{homogeneity} of a segment. This will enables us to take advantage of multiple segmentations, by combining the label confidence of segments based on their homogeneity. We combine the predictions of all three classifiers probabilistically to estimate the final label as described below.
\vspace*{-0.3in}
\paragraph{Training:} We extract the features described in \Cref{sec:features} from each segment of a training video. As the segments vary across time, we opt to extract features for each frame for the same segment to provide discriminating information over time (\eg appearance, motion, and pose of objects) as opposed to sampling features from unique spatio-temporal regions only. In addition, features are extracted independently for different hierarchical segmentation levels to provide instances with more spatial support. We extract features from 5 segmentation hierarchical levels ranging from 10\% to 50\% of the hierarchy height. Segments with a single ground truth label are used to train main and sub-vertical classifiers. We train the homogeneity classifier by providing examples of a single label and ``mix'' label segments as positive and negative instances.
\vspace*{-0.15in}
\paragraph{Prediction:} To predict the labels for a test video, features are extracted from each segment. A spatio-temporal region is labeled on a per-frame basis with the final classification being obtained by averaging the predicted class-posteriors across frames. We label main and sub-vertical geometrical classes independently, \ie we compute the sub-vertical labels for all the segments in a frame but only apply it to segments labeled as vertical by main classifier.

When using multiple segmentations across different hierarchies, a super-pixel is part of different segments at each level of segmentation hierarchy. To determine the label $y_i$ of \emph{super-pixel} $i$, class-posteriors from all segments in the hierarchy $s_j$, containing the super-pixel are combined using a weighted average based on their homogeneity likelihoods $P(s_j|\textbf{x}_j)$~\cite{HoiemIJCV2007,HoiemGeometric2005}, where $\textbf{x}_j$ is the corresponding feature vector. The likelihood of a segment label is then given as:\vspace*{-0.05in}
\[
P(y_i=k|\textbf{x}_i) = \sum\limits_{j}^{n_s} P(y_j = k|\textbf{x}_j,s_j) P(s_j|\textbf{x}_j),
\]
where, $k$ denotes the possible geometric labels and $n_s$ are number of hierarchical segmentations. This technique yields a final classification of super-pixels at the over-segmentation level by combining the individual predictions across hierarchy levels. These weighted posterior probabilities of super-pixels, for main and sub-vertical class, are then averaged across frames in a temporal window to give final predictions for each super-voxel.

\begin{figure*}[ht]
\centering
\includegraphics[width=0.92\textwidth, clip=true, trim=0 1900pt 0 295pt]{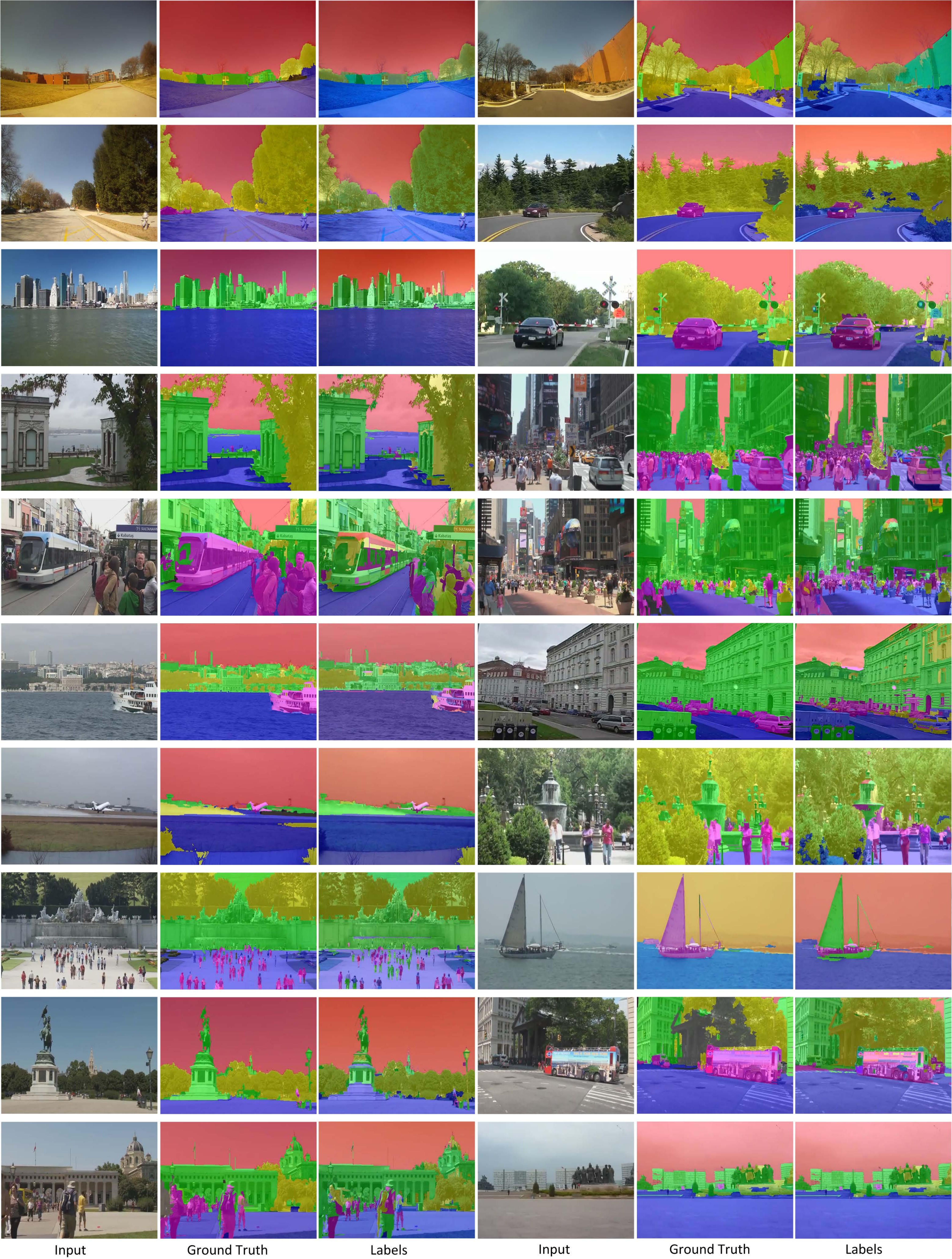}
\setlength\fboxsep{0pt}
\setlength\fboxrule{0.5pt}
\fbox{\includegraphics[scale=.25]{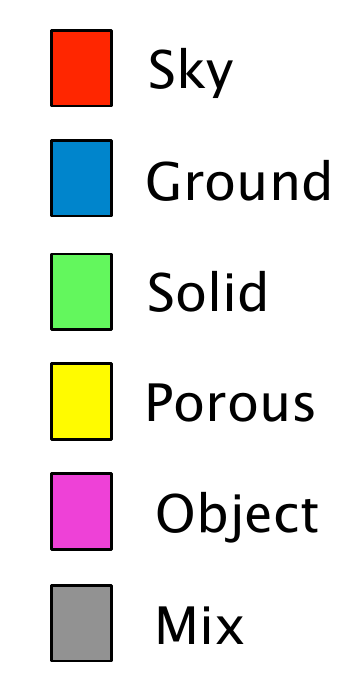}}
\vspace{-0.1in}
\caption{\small{Qualitative results: From left to right: Input video frames, ground truth labels and predicted geometric labels. Our system performs well in challenging settings accurately predicting crowds, objects and foliage.} }
\label{fig:ex}
\end{figure*}

\section{Results}
\label{sec:results}

\noindent We report the accuracy of our method using the number of pixels correctly labeled on the testing videos, \ie a 90\% class accuracy indicates that 90\% of the pixels of that class were labeled correctly. In our experiments, leveraging multiple hierarchy levels and temporal redundancy, we achieve an overall classification accuracy of \emph{96.0\%} for main and \emph{77.4\%} for the sub-vertical classes. After classification, each super-pixel is assigned the probability for each geometric class, as shown in \Cref{fig:vert_conf}. Qualitative results are shown in \Cref{fig:ex}; we encourage the reader to watch the supplementary video.

\begin{table}[hb]
\centering
\begin{subtable}[b]{0.4\textwidth}
\small
\centering
\begin{tabular}{|c|c|c|c|}\hline
&Sky&Ground&Vertical\\\hline
Sky& 99.4  & 0.0  & 0.6   \\\hline
Ground& 1.2  & 96.3  & 2.5   \\\hline
Vertical& 2.9  & 5.1  & 92.0   \\\hline
\end{tabular}
\caption{Main Classes}
\end{subtable}
\begin{subtable}[b]{0.4\textwidth}
\small
\centering
\begin{tabular}{|c|c|c|c|}\hline
&Solid&Porous&Object\\\hline
Solid& 73.8 & 13.0 & 13.2  \\\hline
Porous& 3.4 & 89.2 & 7.4  \\\hline
Object& 11.3 & 19.5 & 69.2  \\\hline
\end{tabular}
\caption{Sub-vertical Classes}
\end{subtable}
\vspace{-0.1in}
\caption{\small{Confusion matrices for main and sub-vertical classfication.}}
\label{Table:confu}
\vspace{-0.1in}
\end{table}

\begin{figure*}[htb]
\centering
\begin{subfigure}[b]{0.33\textwidth}
\includegraphics[scale=.26,trim=10pt 200pt 20pt 250pt]{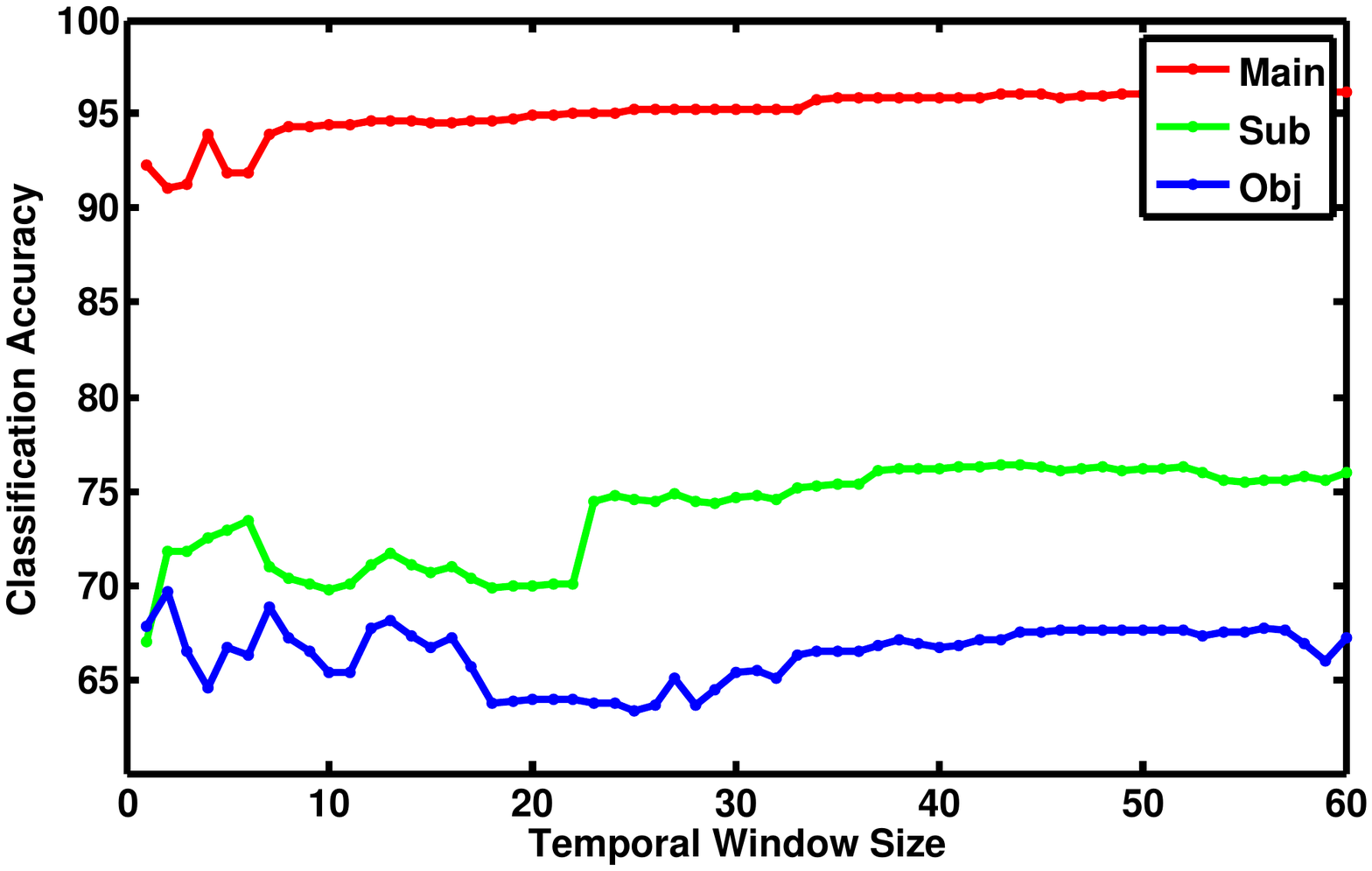}
\caption{Accuracy for different temporal windows.}
\label{fig:temporal_const}
\end{subfigure}
\begin{subfigure}[b]{0.66\textwidth}
\includegraphics[scale=.26, trim=40pt 200pt 30pt 250pt]{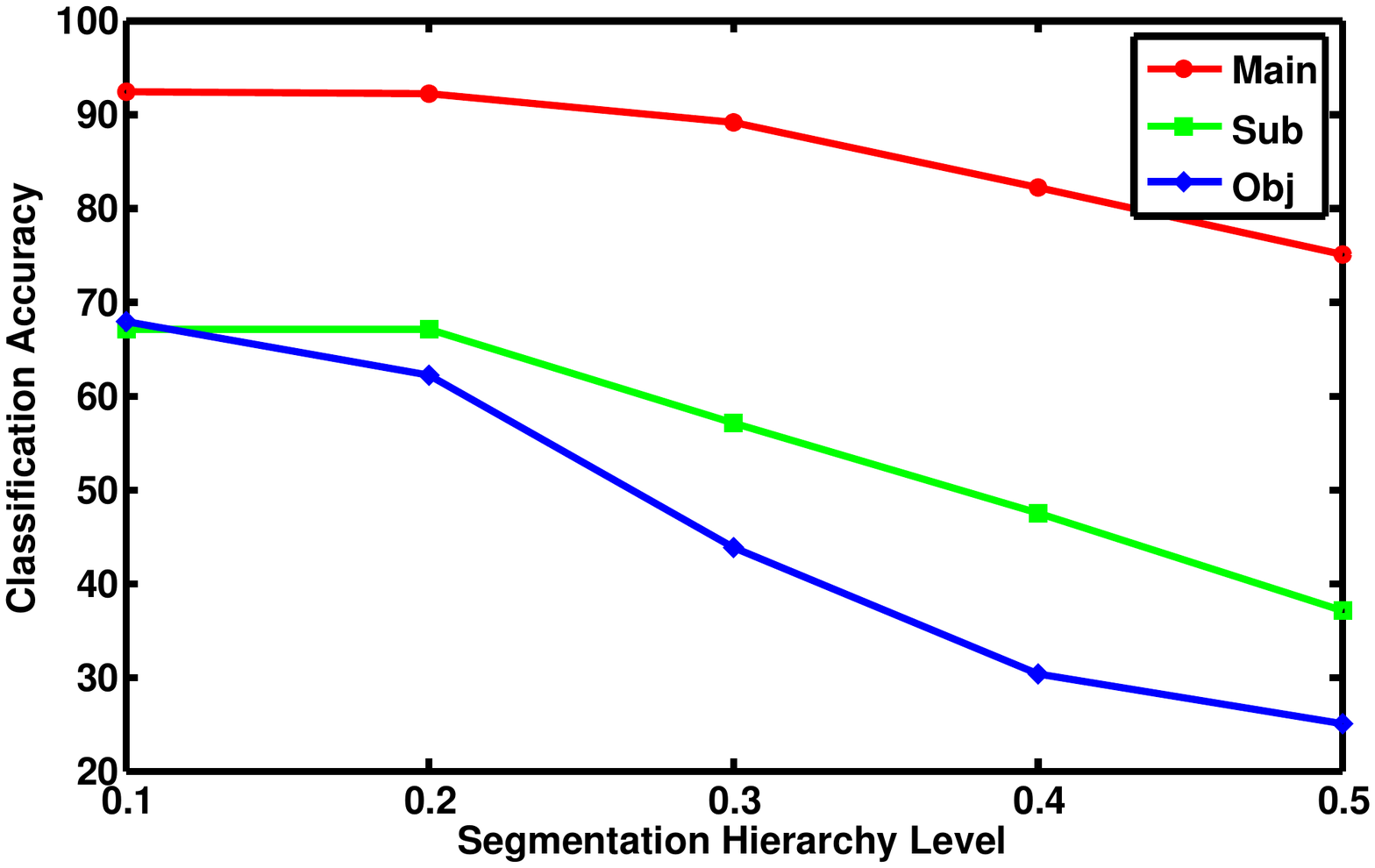}
\includegraphics[scale=.26, trim=0pt 200pt 50pt 250pt]{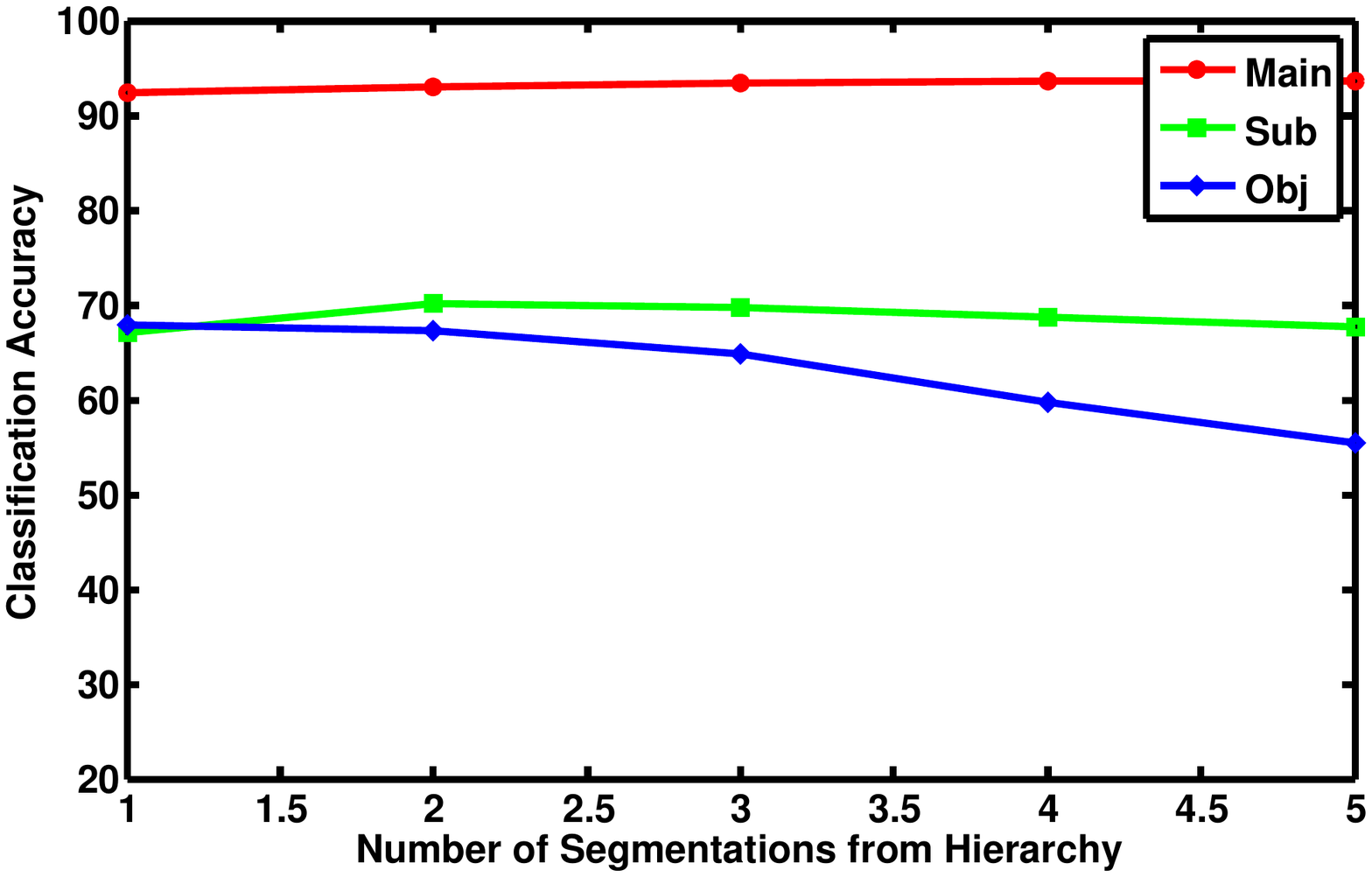}
\caption{Classification results for various hierarchy levels.}
\label{fig:plots}
\end{subfigure}
\vspace{-0.1in}
\caption{\small{(a) Temporal consistency improves accuracy for main and sub-vertical classifcation. The temporal window starts at the very first frame a segment appears in the video. (b) Classification accuracy estimated over 5-fold cross validation: (left) Single segmentation hierarchy level, (right) Multiple segmentation hierarchy levels. A temporal window size of one is used in both experiments.}}
\vspace{-0.1in}
\end{figure*}

It is insightful to quantify to which extent temporal redundancy improves classification accuracy. To this end, we evaluate classification accuracy across different size of temporal windows. Specifically, we compute the class-posteriors of a segment independently for each frame, obtaining the final probability by taking the average of the per-frame probabilities across the temporal window. Figure \ref{fig:temporal_const} shows the result for different lengths of temporal windows. It shows that accuracy reaches a stationary point for temporal windows of size 25 or greater. Using temporal window for labeling improves classification accuracy from 92.3\% for a single frame to 96\% for 25 frames for main classes, and from 67\% to 77.4\% for sub-vertical classes. However, accuracy for objects is virtually unchanged which we believe is due to the difficulty of segmenting these objects.

Figure \ref{fig:plots} demonstrates the variation in classification accuracy when using a single versus multiple segmentation hierarchy levels. When using a single segmentation, the classification accuracy decreases with increasing hierarchy level from 0.1 to 0.5 (here 0.1 denotes the level at 10\% of the overall hierarchy height). This decrease in accuracy is due to segments of different classes being increasingly mixed at higher hierarchy level as regions tend to get under segmented. Using multiple segmentations by combining different segmentation hierarchy levels provides a much more consistent accuracy, in particular it mitigates the problem of determining the correct granularity for a segment. In our experiments, combining predictions for geometric context at hierarchy levels 0.1 and 0.2 yields the best results. 



\begin{figure}[hb]
\centering
\vspace{-0.15in}
\includegraphics[width=0.6\columnwidth]{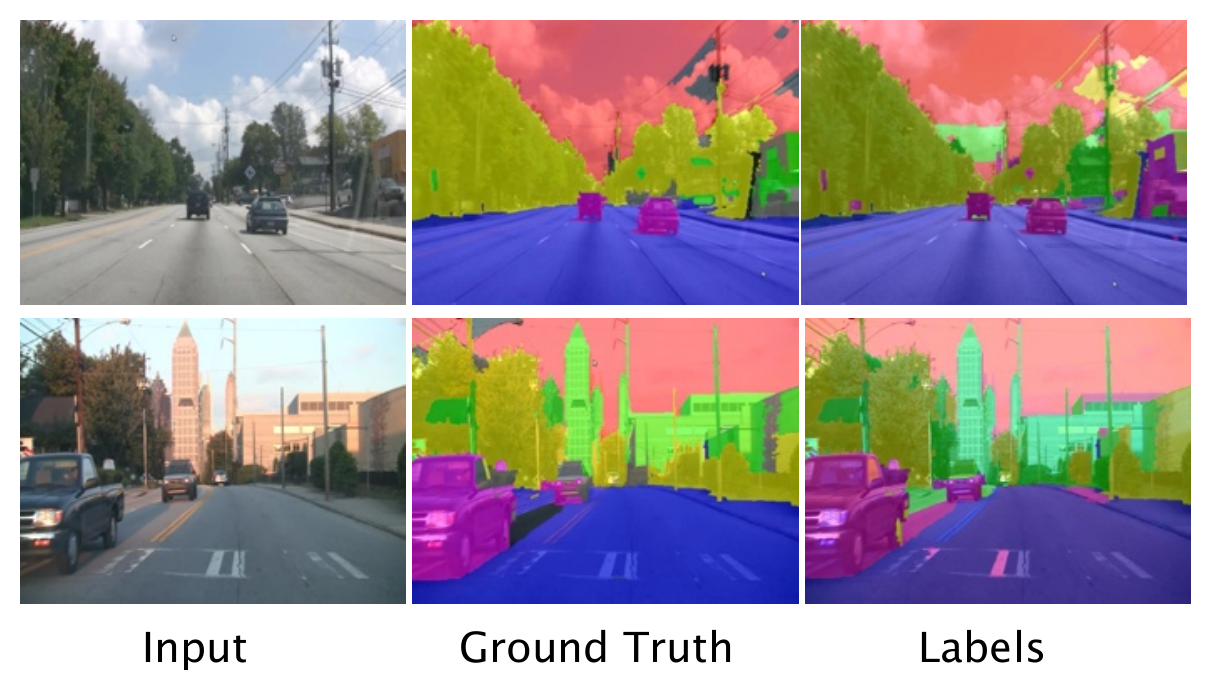}
\vspace{-0.15in}
\caption{\small{Misclassification examples: Scattered clouds are labeled as vertical class, a mix region of object / solid is labeled as car (top). Some ground is mistakenly labeled as object (bottom).}}\vspace{-0.1in}
\vspace{-0.1in}
\label{fig:misclass}
\end{figure}

\Cref{Table:confu} shows the row normalized confusion matrices. Notice, that we are able to achieve highly accurate classification results for main classes. For vertical sub-classes accuracy is lower, due to the vertical class containing huge intra-class variations and regions tend to be more affected by segmentation errors than the other classes. Finally, some qualitative miss-classifications are illustrated in \Cref{fig:misclass}.
\vspace{-0.15in}

\begin{figure}[htb]
\centering
\includegraphics[width=0.8\columnwidth]{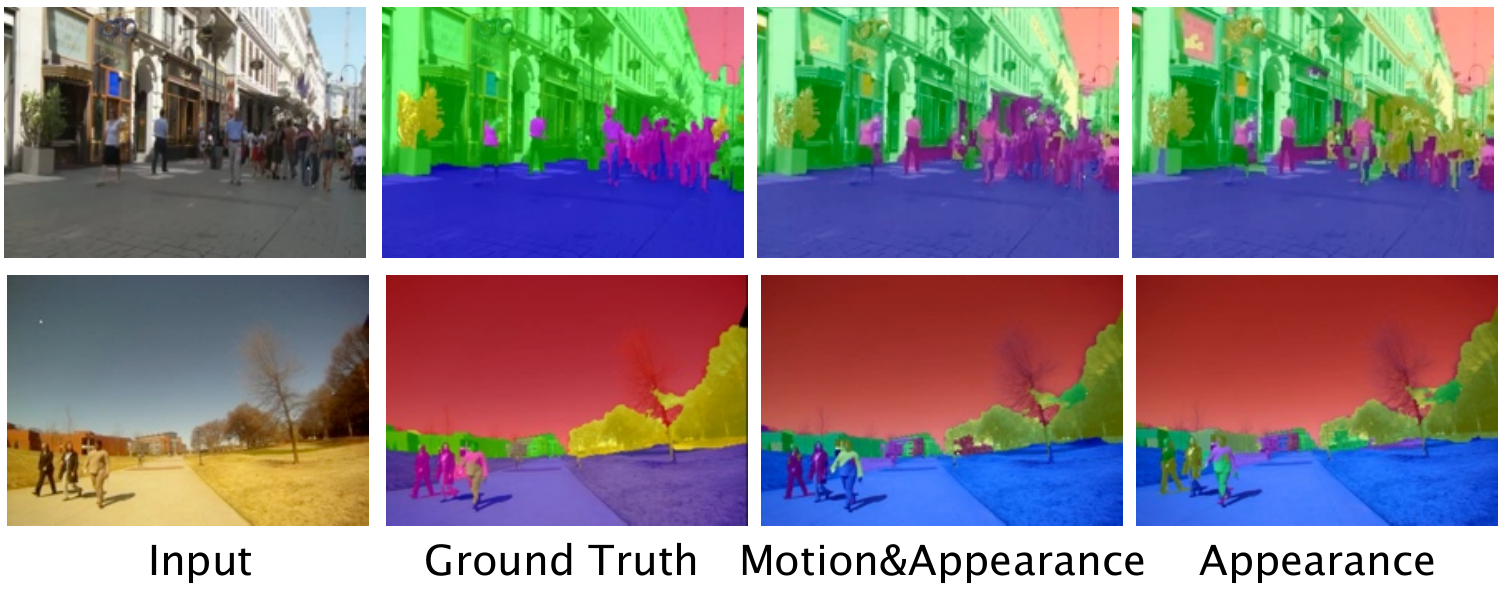}
\includegraphics[width=0.15\columnwidth]{legend}
\vspace{-0.1in}
\caption{\small{Qualitative comparison of importance of features. Overall, motion and appearance features achieve highest accuracy.}}
\vspace*{-0.1in}
\label{fig:feat_imp}
\end{figure}

\paragraph{Importance of Features:}
We are using a wide variety of features covering appearance and motion. Here, we provide some insight into the importance of each individual feature type. 
To estimate the importance of a feature set, we only use the particular feature set across our cross-validation dataset to train and test our system. \Cref{Table:feaImp} shows the difference in accuracy when using only a particular feature set, here for a temporal window size of 1 frame. It can be seen, that the use of motion and appearance features yields the best accuracy, where motion features are primarily beneficial across the sub-vertical classifier (accuracy improves by 5\% compared to appearance features alone).  \Cref{Table:feaImp} also shows the benefit of temporal redundancy by using spatio-temporal regions. Compared to limiting features to only the very first frame for each region (last 2 rows in table),\ie a setting similar to the image case, accuracy increases greatly (by a mean of 9\% on the sub-vertical, and by 69.5\% for the object class, in particular). This change is even more dramatic when comparing using all frames to using only the very first frame if limited to only appearance features (275\%). Qualitative results are shown in \Cref{fig:feat_imp}.

\begin{table}[hb]
\small
\centering
\begin{tabular}{|c|c|c|c|}\hline
\emph{Features}&\emph{Main}&\emph{Sub-Vertical}&\emph{Object}\\\hline
Motion \& Appearance & 92.3  & 67.0  & 67.8     \\\hline
Appearance only& 92.3  & 64.0  & 64.7     \\\hline
Motion only & 87.3  & 52.7  & 57.1 \\\hline
Motion  \& Appearance & & &\\
(first frame of segment only) & 91.1  & 61.4  & 40.0 \\\hline
Appearance (first frame) & 89.6 & 57.8  & 23.5  \\\hline
\end{tabular}
\vspace*{-0.1in}
\caption{\small{Feature importance. We list the mean accuracy for the main and sub-vertical classifier and the individual accuracy of the object classifier. Using motion and appearance features yields the best accuracy (top row). Temporal redundancy is significant to our results, as shown by the reduced accuracy when limiting features to only the very first frame of each segment (last 2 rows).}
}
\label{Table:feaImp}
\vspace*{-0.18in}
\end{table}

\subsection{Semi-supervised Learning}
\label{sec:semisupervised}
\noindent Machine learning systems tend to improve performance with more training data available as intra-class variation is better accounted for. To verify this assumption for our dataset, we randomly pick 12, 24, 36, and 48 videos from the cross-validation dataset and restrict training to this set. \Cref{Table:DataDepend:a} shows that accuracy indeed improves with increasing training data size, verifying our assumption.

\begin{figure*}
\centering
\includegraphics[width=\textwidth]{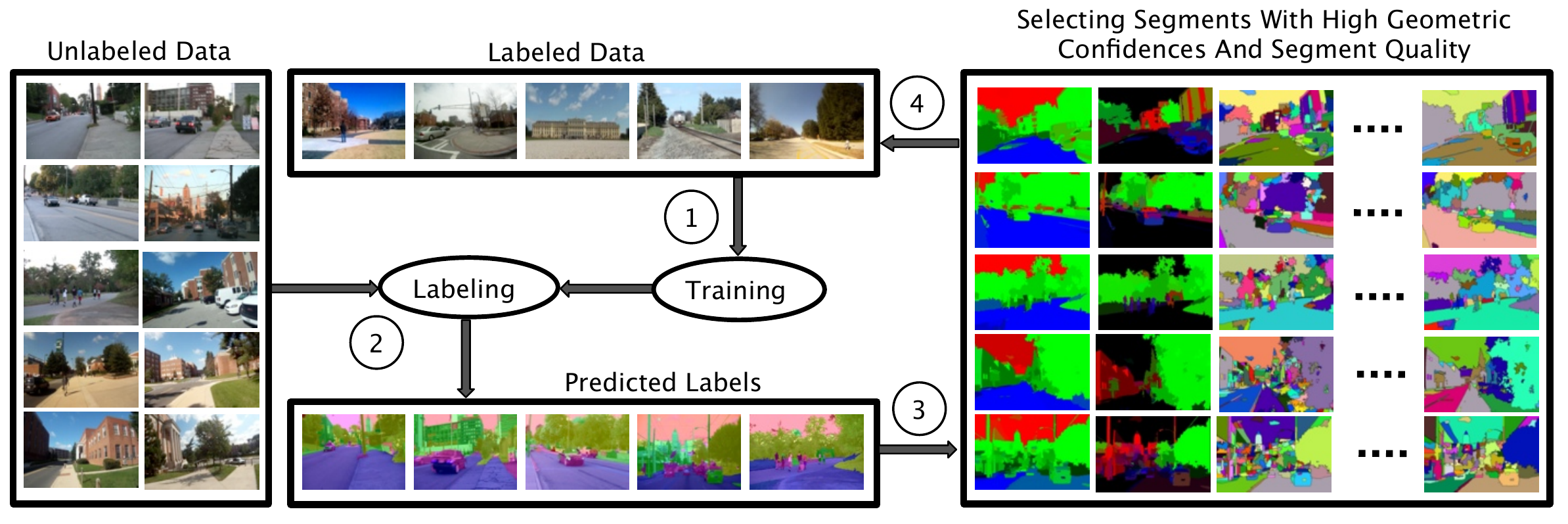}
\vspace*{-0.3in}
\caption{\small{Overview of semi-supervised bootstrap learning for geometric context in video (see \Cref{sec:semisupervised}). 
For segment selection, columns illustrate: (Left) confidence for main-classifier via color intenstiy (vertical: green, sky: red, ground: blue), (2nd) confidence for sub-vertical classifier (solid: red, porous: green, object: blue), (3rd and 4th) Segments across hierarchy levels (randomly colored).}}
\vspace*{-0.1in}
\label{fig:ssl}
\end{figure*}

Obtaining large amounts of training data requires manual annotation of videos, which is time consuming and requires crowd-based approaches \cite{russell2008labelme} to scale. Alternatively, utilizing a large quantity of unlabeled data, we propose to adapt semi-supervised bootstrap learning. We iteratively train classifiers in a self-training manner, as shown in \Cref{fig:ssl}. First classifiers (main, sub-vertical, and homogeneity) are trained using the annotated ground-truth data (1). Then, these classifiers are used to predict geometric context on unlabeled data (2). Segments with most confident labels (maximum class posterior $\ge 80\%$) are selected (3) and added to the training data with their predicted labels (4). In addition, we make use of multiple segmentations at different hierarchy levels, by including all high-confidence segments from the hierarchy that have high homogeneity (probability of being a single class $\ge 80\%$). Finally, the classifiers are re-trained on the expanded pool of labeled data and the process is iterated over. We expect accuracy on the added data to improve over several iterations. To avoid adding low quality segments to the labeled set, we perform introspection every 5th iteration, discarding added segments whose confidence (maximum class posterior) dropped below 80\%.

\Cref{Table:DataDepend} demonstrates the effectiveness of our mutli-segmentation semi-supervised bootstrap learning. Our initial classifier is trained on a dataset of 63 videos (all videos in cross-valiation set, $\sim200,000$ segments). At each iteration, we add 5,000 high confidence segments of each geometric class from unlabeled dataset, extending the training data. After 10 iterations, we are able to improve the performance by 1\% for main, 3\% for subvertical, and 9.5\% for objects. In particular, we evaluate our bootstrap approach on a separate video dataset of 40 videos (7,000 frames). Comparing \Cref{Table:DataDepend:a} and \Cref{Table:DataDepend:b} shows that after 10 iterations we are able to achieve an improvement by semi-supervised bootstrap learning comparable to that of supervised learning.

\begin{table}
\centering
\begin{subtable}[b]{\columnwidth}
\small
\centering
\begin{tabular}{|c|c|c|c|}\hline
\emph{No. of videos}&\emph{Main}&\emph{Sub-Vertical}&\emph{Object}\\\hline
12 & 91.7  & 54.9  & 32.6 \\\hline
24 & 92.4  & 62.1  & 59.3     \\\hline
36 & 92.3  & 66.0  & 65.5     \\\hline
48 & 92.3  & 67.0  & 67.4     \\\hline
\end{tabular}
\caption{Data-size dependency in supervised learning}
\label{Table:DataDepend:a}
\end{subtable}

\begin{subtable}[b]{\columnwidth}
\small
\centering
\begin{tabular}{|c|c|c|c|}\hline
\emph{Iteration}&\emph{Main}&\emph{Sub-Vertical}&\emph{Object}\\\hline
0 & 85.1  & 74.7  & 73.0     \\\hline
5 & 85.2  & 74.2  & 75.0     \\\hline
10 & 86.2  & 77.2 & 79.9     \\\hline
\end{tabular}
\caption{Semi-supervised bootstrap learning}
\label{Table:DataDepend:b}
\end{subtable}
\vspace{-0.2in}
\caption{\small{(a) Accuracy improves with larger training set size in supervised learning setting. (b) Leveraging semi-supervised learning (\Cref{fig:ssl}) accuracy improves with successive iterations.}}  
\label{Table:DataDepend}
\vspace{-0.3in}
\end{table}


\section{Conclusion and Future Work}
\label{conclusion}






\noindent In this paper, we propose a novel algorithm for estimating geometric context in video, achieving highly accurate results. We thoroughly evaluate the contribution of motion features and demonstrated the benefit of utilizing temporal redundancy across frames. To measure accuracy of our approach, we collected a comprehensive dataset of annotated video which we plan to make available to the research community. We further showed how semi-supervised learning can broaden the pool of annotated data. To the best of our knowledge we demonstrate the first temporally consistent results for geometric context on \emph{video}.

In the future, we plan to increase accuracy for the sub-vertical classifier. One reason for its lower accuracy is, that objects tend to be under-segmented even at the superpixel level, merging with porous or solid classes at higher hierarchy levels. We believe that improved segmentation of foreground objects will lead to enhanced accuracy of our method. Finally, we plan on leveraging geometric context to improve object detection and activity recognition in video. 
\vspace{-0.3cm}
\paragraph{Acknowledgement}
\small
This material is based in part on research by the Defense Advanced Research Projects Agency (DARPA) under Contract No. W31P4Q-10-C-0214, and by a Google Grant and and Google PhD Fellowship for Matthias Grundman, who participated in this research as a Graduate Student at Georgia Tech. Any opinions, findings and conclusions or recommendations expressed in this material are those of the authors and do not necessarily reflect the views of any of the sponsors funding this research.




%

{
\footnotesize
\bibliographystyle{plain}
\bibliography{GCFV}
}

\end{document}